\documentclass{article}

\usepackage{microtype}
\usepackage{graphicx}
\usepackage{subfigure}
\usepackage{booktabs} 
\usepackage{paralist}
\usepackage{amssymb}
\usepackage{amsmath}

\usepackage{multirow}

\usepackage[utf8]{inputenc}
\usepackage[english]{babel}
\usepackage{amsthm} 

\usepackage{hyperref}

\usepackage[accepted]{icml2021}

\icmltitlerunning{Connecting Graph Convolutional Networks and Graph-Regularized PCA}

\newcommand{\hide}[1]{}

\newcommand{\bit}{\begin{compactitem}}
\newcommand{\eit}{\end{compactitem}}
\newcommand{\ben}{\begin{compactenum}}
\newcommand{\een}{\end{compactenum}}

\newcommand{\beq}{\begin{equation}}
\newcommand{\eeq}{\end{equation}}

\newcounter{x}

\newcommand{\bz}{\mathbf{z}}

\newcommand{\Z}{{Z}}

\newcommand{\cora}{Cora\xspace}
\newcommand{\seer}{CiteSeer\xspace}
\newcommand{\pub}{PubMed\xspace}
\newcommand{\arxiv}{Arxiv\xspace}
\newcommand{\pro}{Products\xspace}

\newcommand{\mL}{\mathcal{L}}

\newcommand{\tL}{\tilde{L}}
\newcommand{\Lsup}{\tilde{L}_{\text{spr}}}
\newcommand{\tAsup}{\tilde{A}_{\text{spr}}}

\newcommand{\cbit}{\begin{compactitem}}
	\newcommand{\ceit}{\end{compactitem}}
\newcommand{\cben}{\begin{compactenum}}
	\newcommand{\ceen}{\end{compactenum}}

\long\def\ignore#1{}

\DeclareMathOperator{\Tr}{Tr}
\usepackage{xspace}
\newcommand{\method}{{\sc GPCAnet}\xspace}

\newtheorem{theorem}{Theorem}[section]

\begin{document}

\twocolumn[
\icmltitle{Connecting Graph Convolutional Networks and Graph-Regularized PCA}
\icmlsetsymbol{equal}{*}

\begin{icmlauthorlist}
\icmlauthor{Lingxiao Zhao}{cmu}
\icmlauthor{Leman Akoglu}{cmu}
\end{icmlauthorlist}

\icmlaffiliation{cmu}{Carnegie Mellon University, Pittsburgh, PA, USA}
\icmlcorrespondingauthor{Lingxiao Zhao}{lingxia1@andrew.cmu.edu}

\icmlkeywords{Machine Learning, ICML}

\vskip 0.3in
]





\begin{abstract}

Graph convolution operator of the GCN model is originally motivated from a localized first-order approximation of spectral graph convolutions. This work stands on a different view; establishing a \textit{mathematical connection between graph convolution and graph-regularized PCA} (GPCA). 
Based on this connection, GCN architecture, shaped by stacking graph convolution layers, shares a close relationship with stacking GPCA. We empirically demonstrate that the \textit{unsupervised} embeddings by GPCA paired with a 1- or 2-layer MLP achieves similar or even better performance  than GCN on semi-supervised node classification tasks across five datasets including Open Graph Benchmark \footnote{\url{https://ogb.stanford.edu/}}. 
This suggests that the prowess of GCN is driven by graph based regularization.
In addition, we extend GPCA to the (semi-)supervised setting and show that it is equivalent to GPCA on a graph extended with ``ghost'' edges between nodes of the same label.
Finally, 
we capitalize on the discovered relationship to design an effective initialization strategy based on stacking GPCA, enabling GCN to converge faster and achieve robust performance at large number of layers. 
Notably, 
the proposed initialization is general-purpose and applies to other GNNs.

\end{abstract}

\section{Introduction}
  \label{sec:intro}

Graph neural networks (GNNs) are neural networks designed for the graph domain. Since the breakthrough of GCN \cite{kipf2017semi}, which notably improved performance on the semi-supervised node classification problem, many GNN variants have been proposed; including 
GAT \cite{velickovic2018graph}, GraphSAGE \cite{hamilton2017inductive}, DGI \cite{velickovic2018deep}, GIN \cite{xu2018how}, APPNP \cite{klicpera_predict_2019}, to name a few. 

Despite the empirical successes of GNNs in both node-level and graph-level tasks, 
they remain not well understood due to the lack of systematic and theoretical analysis of GNNs. For example, researchers have found that GNNs, unlike their non-graph counterparts, suffer from performance degradation with increasing depth, losing their expressive power exponentially in number of layers \cite{Oono2020Graph}. Such behavior is only partially explained by the oversmoothing phenomenon \citep{li2018deeper, Zhao2020PairNorm:}. Another surprising observation shows that a Simplified Graph Convolution model, named SGC \cite{wu2019simplifying}, can achieve similar performance to various more complex GNNs on a variety of node classification tasks. 
Moreover, a simple baseline that does not utilize the graph structure altogether performs similar to state-of-the-art GNNs on graph classification tasks \cite{Errica2020A}. These observations call attention to studies for a better understanding of GNNs \citep{nt2019revisiting,morris2019weisfeiler,xu2018how,Oono2020Graph,Loukas2020What,Srinivasan2020On}.  (See Sec. \ref{sec:related} for more on understanding GNNs.)

Toward a systematic analysis and better understanding of GNNs, we establish a connection between the graph convolution operator of GCN and Graph-regularized PCA (GPCA) \cite{zhang2012low}, and show the similarity between GCN and stacking GPCA. This connection provides a deeper understanding of GCN's power and limitation. Empirically, we also find that GPCA performance matches that of GCN on benchmark semi-supervised node classification tasks. What is more, the unsupervised stacking GPCA can be viewed as ``unsupervised GCN'' and provides a straightforward, yet systematic way to initialize GCN training. We summarize our contributions as follows:

\cbit
    \item {\bf Connection btwn. Graph Convolution and GPCA:} We establish the connection between the graph convolution operator of GCN and the closed-form solution of graph-regularized PCA formulation. We demonstrate that a simple graph-regularized PCA paired with 1- or 2-layer MLP can achieve similar or even better results than GCN over several benchmark datasets. We further extend GPCA to (semi-)supervised setting which can generate embeddings using information of labels, which yields better performance on 3 out of 5 datasets. The outstanding performance of simple GPCA supports that the prowess of GCN on node classification task comes from graph based regularization. This motivates the study and design of other graph regularizations in the future.
    

    \item {\bf \method: New Stacking GPCA model:~} Capitalizing on the  connection between GPCA and graph convolution, we design a new GNN model called \method shaped by (1) stacking multiple GPCA layers and nonlinear transformations, and (2) fine-tuning end-to-end via supervised training.
    \method is a generalized GCN model with adjustable hyperparameters that control the strength of graph regularization of each layer. We show that with stronger regularization, we can train \method with fewer (1--3) layers and achieve comparable performance to much deeper GCNs.

    \item {\bf First initialization strategy for GNNs:~} Capitalizing on the connection between GCN and \method, we design a new strategy to initialize GCN training based on stacking GPCA, outperforming the popular Xaiver initialization \cite{glorot2010understanding}. We show that the \method-initialization is extremely effective for training deeper GCNs, with greatly improved convergence speed, performance, and robustness. 
    Notably, \method-initialization is general-purpose and also applies to other GNNs.
    To our knowledge, it is the first initialization method specifically designed for GNNs.


    


\ceit

{\bf Reproducibility.~} We open-source all the code developed in this work at \url{http://bit.ly/GPCANet}.
The datasets are already public-domain.


\section{Related Work}
  \label{sec:related}

\textbf{Understanding GNNs.} We discuss related work that provide a more systematic understanding of GNNs in a number of fronts.
GCN's graph convolution is originally motivated from the approximation of graph filters in graph signal processing \cite{kipf2017semi}. NT et al. \yrcite{nt2019revisiting} 
show that graph convolution only performs low-pass filtering on original feature vectors, and also states a connection between graph filter and Laplacian regularized least squares. Motivated by the oversmoothing phenomenon of graph convolution, Oono et al. \yrcite{Oono2020Graph} theoretically prove that GCN can only preserve information of node degrees and connected components when the number of layers goes to infinity, under some conditions of GCN weights. 
At graph-level,
Xu et al. \yrcite{xu2018how} show that GNNs cannot have better expressive power than the Weisfeiler-Lehman (WL) test of graph isomorphism, and develop the GIN model that is as powerful as the WL test. Morris et al. \yrcite{morris2019weisfeiler} extend the work by \cite{xu2018how} and establish a connection to the higher-order WL algorithm. 
When given distinguishable node features, Loukas \yrcite{Loukas2020What} has shown that the GNN models can be Turing universal with sufficient depth and width. The relationship between network capacity and network size is also studied in \cite{loukas2020hard}.
In terms of limitation of expressiveness, Chen et al. \yrcite{chen2020can} studied the attributed graph substructures problem and proved that GNNs cannot count induced subgraphs. Garg et al. \yrcite{garg2020generalization} further showed that message-passing based GNNs cannot compute graph properties such as cycle, diameter and clique informations, also providing a Rademacher generalization bound for binary graph classification. Recently, Liao et al. \yrcite{liao2021a} improved the generalization bound via a PAC-Bayesian approach.

\textbf{Graph-regularized PCA.} 
PCA and its variants are standard linear dimensionality reduction approaches. Several works extend PCA to 
graph-structured data, such as Graph-Laplacian PCA 
 \cite{jiang2013graph} 
and Manifold-regularized Matrix Factorization 
\cite{zhang2012low}. For other variants, see 
 \cite{shahid2016fast}.

\textbf{Stacking Models and Deep Learning.} The connection between CNN and stacking PCA has been explored in PCANet \cite{chan2015pcanet}, 
which demonstrated that the (unsupervised) simple stacking PCA works as good as supervised CNN over a large variety of vision tasks. The original PCANet is shallow and does not have nonlinear transformations, PCANet+ \cite{low2017stacking} overcomes these limitations and pushes the architecture much deeper. The idea of layerwise stacking for feature extraction is not new and was empirically observed to exhibit better representation ability in terms of classification. For a comprehensive review, we refer to \cite{Bengio13}.

\textbf{Initialization.}
Traditionally, neural networks (NNs) were initialized with random weights generated from Gaussian distribution with zero mean and a small standard deviation \cite{krizhevsky2012imagenet}. As training deeper NNs became extremely difficult due to vanishing gradient and activation functions, Glorot et al. \yrcite{glorot2010understanding} provided a specific weight initialization formula, named Xavier initialization, based on variance analysis without considering activation function. Xavier initialization is widely used for any type of NN even today, and it is the main initialization strategy used for GNNs. Later, He et al. \yrcite{he2015delving} adapted Xavier initialization to ReLU activation by considering a multiplier. Taking another direction, Saxe et al. \yrcite{saxe2013exact} analyzed the dynamics of training deep NNs and proposed random orthonormal initialization. Mishkin and Matas \yrcite{mishkin2015all} further improved orthonormal initialization for batch normalization \cite{ioffe2015batch}. Different from these data-independent approaches, others \cite{wagner2013learning,krahenbuhl2015data,seuret2017pca} have employed data-dependent techniques, like PCA, to initialize deep NNs. 
Although initialization has been widely studied for general NNs, no specific initialization has been proposed for GNNs.
In this work, we propose a data-driven initialization technique (based on GPCA), specific to GNNs for the first time.


\section{GPCA and Graph Convolution}
   \label{sec:gpca}

\subsection{Graph Convolution}
Consider a node-attributed input graph $G=(V,E,X)$ with $|V|=n$ nodes and $|E|=m$ edges, where $X\in \mathbb{R}^{n\times d}$ denotes the feature matrix with $d$ features.
Similar to other neural networks stacked with repeated layers, GCN \cite{kipf2017semi} contains multiple graph convolution layers each of which is followed by a nonlinear activation. Let $H^{(l)}$ be the $l$-th hidden layer  representation, then GCN follows:
\begin{equation}\label{eq:gcn}
H^{(l+1)} = \sigma(\tilde{A}_{\text{sym}}H^{(l)}W^{(l)})
\end{equation}
where $\tilde{A}_{\text{sym}}= \tilde{D}^{-\frac{1}{2}}(A+I)\tilde{D}^{-\frac{1}{2}}$ denotes the $n\times n$ symmetrically normalized adjacency matrix with self-loops, 
$\tilde{D}$ is the diagonal degree matrix where $\tilde{D}_{ii} = 1+\sum_{j=1}^n A_{ij}$,
$W^{(l)}$ is the $l$-th layer parameter (to be learned), and $\sigma$ is the nonlinear activation function. 

The graph convolution operation is defined as the formulation before activation in Eq. \eqref{eq:gcn}. Formally, graph convolution is parameterized with $W$ and maps an input $X$ to a new representation $\Z$  as

\vspace{-0.2in}
\begin{equation}\label{eq:gconv}
\Z = \tilde{A}_{\text{sym}}XW \;.
\end{equation}

\subsection{Graph-regularized PCA (GPCA)}
As stated by \cite{Bengio13}, ``Although depth is an important part of the story, many \textit{other priors} are interesting and can be conveniently captured 
when the problem is  cast as one of learning a representation.''
GPCA is one such representation learning technique with a graph-based prior.

Standard PCA learns $k$-dimensional projections $\Z\in \mathbb{R}^{n\times k}$ of feature matrix  $X\in \mathbb{R}^{n\times d}$, aiming to minimize the reconstruction error 
\beq
\label{eq:pcaobj}
\Vert X-\Z W^T\Vert_F^2\;,
\eeq
subject to $W\in \mathbb{R}^{d\times k}$ being an orthonormal basis. 
GPCA extends this formalism to graph-structured data by additionally assuming either smoothing bases \cite{jiang2013graph} or smoothing projections \cite{zhang2012low} over the graph. In this work we consider the latter case where low-dimensional projections are smooth over the input graph $G$, with its 
normalized Laplacian matrix 
$\tilde{L} = I - \tilde{A}_{\text{sym}}$. The objective formulation of GPCA is then given as
\begin{align}
\min_{\Z,W} \quad & \Vert X-\Z W^T\Vert_F^2 + \alpha \Tr(\Z^T \tilde{L} \Z) \label{eq:gpca} \\
\text{s.t.} \quad & W^TW = I \label{eq:gpcaconstraint}
\end{align}
where $\alpha$ is a hyperparameter that balances reconstruction error and the variation of the projections over the graph. Note that the first part of Eq. \eqref{eq:gpca}, along with the constraint, corresponds to the objective of the original PCA, while the second part is a
graph regularization term that aims to ``smooth'' the new representations $\Z$ over the graph structure. As such, GPCA becomes the standard PCA when $\alpha=0$. 

Similar to PCA, the problem (\ref{eq:gpca}-\ref{eq:gpcaconstraint}) is non-convex but has a closed-form solution \cite{zhang2012low}. Surprisingly, as we show, it has a close connection with the graph convolution formulation in Eq. \eqref{eq:gconv}. In the following, we give the GPCA solution and then detail its connection to graph convolution in the next subsection.

\begin{theorem}\label{thm:unsupgpca}
GPCA with formulation shown in (\ref{eq:gpca}-\ref{eq:gpcaconstraint}) has the optimal solution $(Z^*, W^*)$ following
\begin{align*}
    W^* &= (\mathbf{w}_1, \mathbf{w}_2,..., \mathbf{w}_k) \\
    Z^* &= (I+\alpha \tilde{L})^{-1} X W^*
\end{align*}
where $\mathbf{w}_1, \mathbf{w}_2,..., \mathbf{w}_k$ are the 
eigenvectors corresponding to the largest $k$ eigenvalues of the matrix $X^T(I+\alpha \tilde{L})^{-1}X$.
\end{theorem}

\begin{proof}
    We give the proof in two steps.
    
    \textit{Step 1: For a fixed $W$, Solve optimal $\Z^*$ as a function of $W$:}~ When fixing $W$ as constant, the problem becomes quadratic and convex.  There is a unique solution, given by first-order optimal condition. Let $\ell$ denote the objective function as given in \eqref{eq:gpca}. Its  gradient can be calculated as
    \begin{align}
    \label{Yder}
    {\partial \ell \over \partial \Z} = 2(I+\alpha \tilde{L})\Z - 2XW \;.
    \end{align}
    Setting \eqref{Yder} to $0$ leads to the solution $\Z^* = (I+\alpha \tilde{L})^{-1} X W$.
    
    \textit{Step 2: Replace $\Z$ with $\Z^*$, Solve optimal $W^*$:}~
    Substituting $\Z$ in objective $\ell$ with $\Z^* = (I+\alpha \tilde{L})^{-1} X W$, we reduce the optimization to
    \begin{multline}
        \label{eq:objplug}
    \min_{W, W^TW= I}  \;\; \Vert X-(I+\alpha \tilde{L})^{-1} X WW^T\Vert_F^2 \;+\; \\\alpha \Tr\big[W^TX^T (I+\alpha \tilde{L})^{-1} \tilde{L} (I+\alpha \tilde{L})^{-1} X W\big]     \;.
    \end{multline}
For this part only, let $M=(I+\alpha \tilde{L})^{-1}$ to simplify the notation. We can show that \eqref{eq:objplug} is equivalent to
\begin{align}\label{eq:gpcaa}
 & \min_{W, W^TW= I}  \quad  \Tr(XX^T + MXWW^TWW^TX^TM) - \nonumber 
 \\ 
 & 2\Tr(MXWW^TX^T) + \alpha \Tr (  W^T X^T M \tilde{L} M XW )
 \end{align}
 Using the cyclic property of (Tr)ace (and plugging $(I+\alpha \tilde{L})^{-1}$ for $M$ back), we can write it as (see Supp. \ref{derivation:trace} for detailed derivation.)
 \begin{align}\label{eq:gpca-eig}
    \max_{W, W^TW= I} \quad &\Tr \big[  W^TX^T(I+\alpha \tilde{L})^{-1} XW \big] \;.
\end{align}
Based on the spectral theorem of PSD matrices, the optimal solution $W^*$ of problem \eqref{eq:gpca-eig} is the combination of eigenvectors, associated with the largest $c$ eigenvalues of the graph-revised covariance matrix  $X^T(I+\alpha \tilde{L})^{-1} X$.
\end{proof}

\subsection{Connection btwn. Graph Convolution and GPCA}
Let $\Phi_{\alpha}:= I + \alpha \tilde{L}$. 
The normalized Laplacian matrix $\tilde{L}$ has absolute eigenvalues bounded by 1, thus, all its positive powers have bounded operator norm.
When $\alpha \le 1$, $\Phi_{\alpha}^{-1}$ can be decomposed into Taylor series as
\begin{align}\label{eq:inverse_poly}
( I + \alpha \tilde{L})^{-1} = I - \alpha \tilde{L} +...+ (-\alpha)^t \tilde{L}^t + ...
\end{align}
The first-order truncated form (i.e. approx.) of Eq. \eqref{eq:inverse_poly} is 
\begin{align}\label{eq:inverse_truncated}
( I + \alpha \tilde{L})^{-1} \approx I - \alpha \tilde{L} = (1-\alpha)I + \alpha \tilde{A}_{\text{sym}} \;.
\end{align}
When $\alpha=1$, the first-order approximation of $\Z^*$ follows 
\begin{align}
\label{eq:approxgpca}
\Z^* \approx  \tilde{A}_{\text{sym}} X W^* \;.
\end{align}
The (approximate) solution to GPCA in Eq. \eqref{eq:approxgpca} matches the graph convolution operation in Eq. \eqref{eq:gconv}, with $W^*$ plugged in as the eigenvectors of the matrix $X^T\Phi_{\alpha}^{-1} X$. 

We restate the key contribution of this paper: {\centering \textit{
The graph convolution operation can be viewed as the first-order approximation of GPCA with $\alpha=1$ with a learnable $W$. Put differently, the first-order approximation of (unsupervised) GPCA with $\alpha=1$ can be viewed as a graph convolution with a fixed, data-driven $W$.}
Notably, for $\alpha < 1$, Eq. \eqref{eq:inverse_truncated} shows the connection between GPCA and graph convolution equipped with 1-step (scaled) residual connection.}

\subsection{Supervised GPCA}
\label{sec:sgpca}

The standard GPCA problem as given in (\ref{eq:gpca}-\ref{eq:gpcaconstraint}) is unsupervised. In this section, we show how to extend it to the supervised setting. 
Here, besides (1) providing good reconstruction and (2) varying smoothly over the graph structure, we would want to learn embeddings that also (3) highly correlate with the response, or outcome variable(s).
For simplicity of presentation, let $\bz \in \mathbb{R}^d$ be a 1-d embedding and $Y$ denote the response matrix (considering the general case of multiple responses). 
We write the additional objective as
\beq
\label{eq:corr}
\max_{\bz} \quad\;  \big[\text{corr}(Y, \bz)  \big]^T \big[\text{corr}(Y, \bz)  \big] \; \text{var}(\bz) 
\eeq
which aims to maximize the correlation between $\bz$ and $Y$.\footnote{Note that for the optimization to be well-posed, constraints on $\bz$ would be required which we omit for simplicity of presentation.} The form of \eqref{eq:corr} and the variance-maximizing term $\text{var}(\bz)$ are for mathematical convenience, that will become explicit in the following. 
Despite agnostic to labels,
 including $\text{var}(\bz)$ is intuitive since an implicit objective of data projection (embedding) is to ensure that inherent variation (structure) in data is captured as much as possible. (Recall from vanilla PCA that the objective of minimizing reconstruction error in \eqref{eq:pcaobj} is equivalent to maximizing the variance of data in the projected space.)

We can write \eqref{eq:corr} in solely matrix form, through a series of transformations (See Supp. \ref{derivation:var}), as
\begin{align}
\max_{\bz} \quad \; &  \big[\text{corr}(Y, \bz)  \big]^T \big[\text{corr}(Y, \bz)  \big]  \text{var}(\bz) \nonumber\\
\equiv \max_{\bz} \quad \;&  \bz^T Y Y^T \bz
\end{align}
In the general case, we would aim to maximize the trace of $\Z^T Y Y^T \Z$
for multi-dimensional embeddings.

{\bf Interpretation.~} 
For semi-supervised node classification with $c$ classes, let $\mL \subset V$ denote the set of labeled nodes. 
For this task, $Y\in \{0,1\}^{n\times c}$ would encode the node labels where the $v$-th row of $Y$, denoted $Y_v$, depicts the one-hot encoded label for each $v\in \mL$. For 
$u\in V\backslash \mL$ with unknown labels, $Y_u=\mathbf{0}$, set as the $c$-dimensional all-zero vector. 
 
Then, $(YY^T)_{ij}$  is simply equal to 1 when nodes $i$ and $j$ share the same label, and otherwise 0 (either because they have different labels or labels are unknown, i.e. $Y_i$ and/or $Y_j$ are all 0's).
This term simply enforces the representations $Z_i$ and $Z_j$ of two same-labeled nodes to be similar. 
In a sense, $YY^T$ is adding ``ghost'' edges between the same-label nodes, further guiding the smoothness of their representations over this extended graph structure. This is particularly beneficial when graph smoothing may not be enough to ensure two nodes of the same label  have similar embeddings when
 they are not directly connected or are far away in the graph. 

We remark that earlier work \cite{conf/kdd/GallagherTEF08} have heuristically introduced edges between same-label nodes to enhance a given graph for the node classification task. In this work, we have derived the theoretical underpinning for this strategy.

{\bf Supervised formulation.~} 
We have shown that requiring the embeddings to correlate with known labels can be interpreted as additional smoothing over ``ghost'' edges between the same-label nodes in the graph.
As such, we extend the GPCA problem (\ref{eq:gpca}-\ref{eq:gpcaconstraint}) to the (semi-)supervised setting as 
\begin{align}
\min_{\Z,W} \quad &  \Vert X- \Z W^T\Vert_F^2  + \alpha \Tr(\Z^T \Lsup \Z)  \label{eq:supgpca}  \\
\qquad \text{s.t.}  \quad & W^TW = I \label{eq:supgpcaconstraint} \;\;\;; \\
\text{where}  \;  \Lsup  & = I - \tAsup \nonumber \\
\quad \tAsup & = (1-\beta) \tilde{A}_{\text{sym}} + \beta {D}^{-\frac{1}{2}}(  YY^T ) {D}^{-\frac{1}{2}} \label{eq:beta}
\end{align}
In Eq. \eqref{eq:beta}, $\beta$ is an additional hyperparameter for trading-off the graph-based regularization (i.e. smoothing) due to the actual input graph edges versus the ones introduced through $YY^T$ between the nodes of the same label, 
and ${D}$ is the diagonal matrix with ${D}_{ii} = \sum_{j=1}^n (YY^T)_{ij}$.


\begin{theorem}
    \label{thm:supgpca}
    Supervised GPCA with formulation shown in (\ref{eq:supgpca}-\ref{eq:supgpcaconstraint}) has the optimal solution $(\Z^*, W^*)$ following
    \begin{align}
    W^* &= (\mathbf{w}_1, \mathbf{w}_2,..., \mathbf{w}_k) \\
    \Z^* & = (I+\alpha \tilde{L}_{\textit{\em spr}})^{-1} X W^* \label{eq:supzsol}
    \end{align}
    where $\mathbf{w}_1,\ldots, \mathbf{w}_k$ are the top
    eigenvectors of matrix 
    $X^T(I+\alpha  \tilde{L}_{\textit{\em spr}})^{-1}X$, equivalently
    $X^T \big( (1+\alpha) I -  \big[  \alpha (1-\beta) \tilde{A}_{\text{sym}} + \alpha\beta{D}^{-\frac{1}{2}}  YY^T {D}^{-\frac{1}{2}}  \big]   \big)^{-1}X$,
    corresponding to the largest $k$ eigenvalues.
\end{theorem}
\begin{proof}
    The proof is similar to that of Theorem \ref{thm:unsupgpca}.
\end{proof} 

The first-order approximation of the matrix inverse in Eq. \eqref{eq:supzsol}
can be written as $I- \alpha \Lsup = (1-\alpha) I + \alpha  
\tAsup$. 
 As such, when $\alpha=1$,
\hspace{-0.15in} 
\begin{align}
\Z^* \approx \big[ (1-\beta) \tilde{A}_{\text{sym}} + \beta {D}^{-\frac{1}{2}}(  YY^T ) {D}^{-\frac{1}{2}} \big]  X W^*\;.
\end{align}
Similar to Eq. \eqref{eq:approxgpca}, the first-order approximation of the supervised GPCA with $\alpha=1$ can still be viewed as a graph convolution, this time on a graph enhanced with what-we-called ``ghost'' edges.

\subsection{Approximation and Complexity Analysis}\label{sec:approx}

According to formulations in Theorems \ref{thm:unsupgpca} and \ref{thm:supgpca}, obtaining 
$W^*$$\in$$\mathbb{R}^{d\times k}$ and 
$\Z^*$$\in$$\mathbb{R}^{n\times k}$ requires two demanding computations
(1) the inverse of $\Phi_{\alpha} = (I+\alpha \tL) \in \mathbb{R}^{n\times n}$, or in the supervised case $\Phi_{\alpha} = (I+\alpha \Lsup)$; and (2) top $k$ eigenvectors of the matrix ${X}^T\Phi_{\alpha}^{-1} X \in \mathbb{R}^{d\times d}$. 
Eigen-decomposition takes $O(d^3)$ \cite{pan1999complexity}, which is scalable as $d$ is usually small. 
Computing matrix inverse, on the other hand, can take $O(n^3)$ and require $O(n^2)$ memory, 
which would be infeasible for very large graphs.

To reduce computation and memory
complexity, we instead approximately compute $F := \phi_{\alpha}^{-1}X$, which is what we really need for both $W^*$ and $\Z^*$. We can equivalently write
{\small{
\begin{align}
(I+\alpha \tL) F = X
\implies & F + \alpha F  =   \alpha P F + X \nonumber \\
\implies & F = \frac{\alpha}{1+\alpha} P F + \frac{1}{1+\alpha}X  \nonumber 
\end{align}
}}
\noindent
where $P=\tilde{A}_{\text{sym}}$ in the unsupervised case and $P = (1-\beta) \tilde{A}_{\text{sym}} + \beta {D}^{-\frac{1}{2}}(  YY^T ) {D}^{-\frac{1}{2}}$ when supervised. 

Then, we can iteratively (with total $T$ iterations) use the power method \cite{matrixmethods89} to compute $F$ as
\begin{align}\label{eq:power}
F^{(t+1)}  \leftarrow   \frac{\alpha}{1+\alpha} P F^{(t)} + \frac{1}{1+\alpha} X 
\end{align}
where $t \in \{0, ..., T\}$ depicts the iteration and $F^{(0)} \in \mathbb{R}^{n\times d}$ is initialized as $X$ (or randomly).
For the supervised case, $P F^{(t)}$ is computed through a series of (from right to left) matrix-matrix products.
This avoids the explicit construction of matrix $YY^T$ in memory.
 Overall, solving for $F$ takes $O(T(m+n)d)$ where $m$ is the number of edges in the graph. The supervised case has an additional term $O(Td|\mL|c)$ with $c$ being the number of classes and $|\mL| \le n$ be the number of labeled nodes, which can also be upper-bounded by $O(T(m+n)d)$ when treating $c$ as constant. 

 Having solved for $F$, we perform the matrix-matrix product  $Z^*=FW^*$ in $O(ndk)$ and then the eigen-decomposition of $X^TF$ in $O(d^3+nd^2)=O(nd^2)$ ($n\ge d$). Assuming $O(d) =O(k)$, overall complexity for computing single layer GPCA is given as $O(Tmd + Tnd + nd^2)$, which is \textit{linear in the number of nodes and edges}. Note that empirically we found $5\le T\le 10$ to be sufficient.

\section{\method: A Stacking GPCA Model}
    \label{sec:gpcanet}

\subsection{\method}
Thus far, we drew a connection between the geometrically motivated manifold-learning based GPCA and the graph convolution operation of deep neural network based GCN. Next we capitalize on this connection to 
design a new model called \method that takes advantage of the relative strengths of each paradigm; namely, 
GPCA's ability to capture data variation and structure (i.e. data manifold), and GCN's ability to capture multiple levels of abstraction (high-level concepts) through stacked layers and non-linearity.

In a nustshell, \method is a stacking of multiple (unsupervised or supervised) GPCA 
layers and nonlinear transformations, which shares the same architecture as a multi-layer GCN.
It consists of two main stages: (1) \textit{Pre-training}, which sets the layer-wise parameters through closed-form GPCA solutions, and (2) \textit{End-to-end-training}, which refines the parameters through end-to-end gradient-based minimization of a global supervised loss criterion at the output layer. 

We remark that \method is \textit{not} the same as GCN, as each convolution layer uses the formulation in Theorems \ref{thm:unsupgpca} and \ref{thm:supgpca} 
(with approximation shown in Sec. \ref{sec:approx}). In fact, when $\alpha=1$ and $\beta=0$, \method is the GCN model initialized with \method-initialization, which we discuss more in Sec. \ref{sec:init}. In other words, \method is a \textit{generalized} GCN model with additional hyperparameters, $\alpha$ and $\beta$, controlling the strength of graph regularization based on the existing or ``ghost'' edges, respectively. 


\subsubsection{Forward Pass and Pre-training stage}

During pre-training, weights of the $l$-th layer, denoted as $W^{(l)} \in \mathbb{R}^{d_{l-1}\times d_{l}}$, are pre-set (i.e. initialized) as the leading $d_{l}$ eigenvectors of the matrix ${H^{(l-1)}}^T\Phi_{\alpha}^{-1} H^{(l-1)}$,\footnote{If $d^{(l)}$ is greater than the number of eigenvectors, all eigenvectors are used, with additional vectors generated from random projection of eigenvectors.} where $H^{(l-1)}$ is the representation as output by the $(l-1)$-th layer (with $H^{(0)}:=X$), and $\Phi_{\alpha}$ can be the unsupervised $(I+\alpha \tL)$ or the supervised $(I+\alpha \Lsup)$. 

The pre-training stage takes a single forward pass. Alg. \ref{alg:GPCANet} shows both the forward pass of \method used during end-to-end-training stage and the procedure of pre-training.
Note that {\color{blue} the line marked in blue} is an additional step used only for pre-training. 
\begin{algorithm}[!t]
  \caption{\textbf{\method Forward Pass and Pre-training}}
  \label{alg:GPCANet}
\begin{algorithmic}[1]
	\small{
  \STATE {\bfseries Input:} graph $G=(V,E,X)$, 
  GPCA hyper-param.(s) $\alpha$ (and $\beta$, if supervised), 
  \#layers $L$, hidden sizes $\{d_1,\ldots,d_L\}$, activation func. $\sigma(\cdot)$, \#apprx. steps $T$
  \STATE {\bfseries Output:} pre-set layer-wise param.s $\{W^{(1)},\ldots,W^{(L)}\}$
  \STATE Initialize $H^{(0)} := X$.
  \FOR{$l=1$ {\bfseries to} $L$}
    \STATE Center  $H^{(l-1)}$ by subtracting mean of row vectors 
    \STATE $F\xleftarrow[]{} H^{(l-1)}$ 
    \FOR{$t=1$ {\bfseries to} $T$}
        \STATE $PF\xleftarrow[]{} (1-\beta) \tilde{A}_{\text{sym}}F + \beta {D}^{-\frac{1}{2}}(  YY^T ) {D}^{-\frac{1}{2}}F$ 
        \STATE $F\xleftarrow[]{} \frac{\alpha}{1+\alpha} PF + \frac{1}{1+\alpha} H^{(l-1)} $ 
    \ENDFOR   
    {\color{blue} \STATE $W^{(l)} \xleftarrow[]{}$  top $d_l$ eigenvectors of ${H^{(l-1)}}^T F$ }
    \STATE $H^{(l)} \xleftarrow[]{} \sigma(FW^{(l)})$ 
  \ENDFOR
}
\end{algorithmic}
\end{algorithm}
\setlength{\textfloatsep}{0.00in}

\textbf{Additional treatment for ReLU:}  
Transformations like ReLU improves model capacity of \method as it can capture nonlinear representations. However at pre-training stage, it causes information loss as all negative values 
are truncated to 0, which hinders the advantage of using the leading $d_l$  
eigenvectors to initialize the weights so as to convey maximum variance (i.e. information) to next layers. To address this issue, 
we instead use the leading $d_l/2$ eigenvectors $\{e_i\}_{i=1}^{d_l/2}$ and their negatives $\{-e_i\}_{i=1}^{d_l/2}$ to initialize $W^{(l)}$. Empirically we observe this always improves performance when using ReLU activation. 

\subsubsection{End-to-end-training stage}
\vspace{-0.05in}

Pre-training can 
be seen as an information-preserving initialization, as compared to an uninformative random initialization),
after which we refine the layer-wise parameters via gradient-based optimization w.r.t. a supervised loss criterion at the output layer. Specifically for semi-supervised node classification, we perform an end-to-end training w.r.t. the cross-entropy loss on the labeled nodes. All parameters are updated jointly through backpropagation during this stage, with forward computation shown in Alg.\ref{alg:GPCANet}.

\subsection{\method-initialization for GCN}\label{sec:init}
When $\alpha=1$, $\beta=0$, and approximating matrix inverse $(I+\alpha\tL)^{-1}$ via first-order truncated Taylor expansion shown in Eq. \eqref{eq:inverse_truncated} , \method has the same architecture with GCN. As such, we can use the pre-training stage of \method to initialize GCN with only minor modification. Specifically,  we replace lines 6 through 10 in Alg. \ref{alg:GPCANet} with a single line:
\begin{align*}
 F \xleftarrow[]{} \tilde{A}_{\text{sym}}H^{(l-1)}
\end{align*} 
Although the modified initialization is for GCN, and is driven by the mathematical connection between \method and GCN that we established, 
 \method-initialization can be used as general-purpose, for other GNNs as well.

\section{Experiments}
    \label{sec:exp}

\vspace{-0.05in}
In this section we design extensive experiments to answer the following questions.

\vspace{-0.05in}
\cbit
\item 
How does the parameter-free, single-layer GPCA compare to layer-wise parameterized, multi-layer GCN?

\item To what extent is {supervised} GPCA beneficial?

\item 
As \method generalizes GCN, can it outperform GCN with the ability to tune regularization via its
 additional hyperparameters $\alpha$ and $\beta$?


\item 
Does \method-initialization help us train better GCN models, in terms of 
accuracy and robustness, providing better generalization especially with increasing model size (i.e. depth)? 

\ceit
\vspace{-0.05in}


\subsection{Experimental Setup}
\vspace{-0.05in}
{\bf Datasets.~} We focus on the semi-supervised node classification (SSNC) problem and use 5 benchmark datasets: 
The first 3 datasets, \cora, \seer, \pub \cite{sen2008collective}, are relatively small (2K to 10K nodes) but widely-used citation networks. For these we use the same data splits as in \cite{kipf2017semi}.
The others, \arxiv and \pro, are newest and larger (100K to 2000K) node classification benchmarks from Open Graph Benchmark \cite{hu2020open}, for which we use the official data splits based on real-world scenarios with potential distribution shift. Data statistics can be found in Supp. \ref{apdx:data}. 

\vspace{-0.05in}
{\bf Baseline.~} 
For baseline, we only use GCN, as experiments are conducted to verify the established connection between GCN and GPCA instead of achieving the state-of-the-art performance on SSNC. Our analysis also provides insights toward a better understanding of GNNs.

\vspace{-0.05in}
{\bf Model configuration.~} 
For hyperparameters (HPs), we define a separate pool of values for hidden size, number of layers, weight decay, dropout rate, and regularization trade-off terms $\alpha, \beta$ for each dataset, where all methods share the same HP pools. Models are trained on every configuration across HP pools and picked based on validation performance. We use the Adam optimizer for all models. Learning rate is first manually tuned for each dataset to achieve stable training, and the same learning rate is fixed for all models---we empirically observed that learning rate is sensitive to datasets but insensitive to models. For GPCA and \method, number of power iterations in Eq. \eqref{eq:power} is always set to 5. 
All experiments use the maximum training epoch as $1000$ and repeat 5 times. Detailed configuration of HPs can be found in Supp. \ref{apdx:hps}. We mainly use a single GTX-1080ti GPU for small datasets \cora, \seer, and \pub. RTX-3090 GPU is used for \arxiv and \pro.

\vspace{-0.05in}
\textbf{Mini-batch training.~} As nodes are not independent, GNN is mostly trained in full-batch under semi-supervised setting. We use full-batch training for all datasets except \pro, which is too large to fit into GPU memory during training. ClusterGCN \cite{chiang2019cluster}, a subgraph based mini-batch training algorithm, is used to train GCN and \method. For evaluation, we still use full-batch since a single forward pass can be conducted without memory issues. Initialization is also employed in full-batch.

\vspace{-0.05in}
\textbf{Fair evaluation.~} Instead of picking the hyperparameter configurations manually, 
every value reported in the following sections is based on the \textit{best} configuration selected using validation performance, where all models leverage the \textit{same} hyperparameter pools. 
Further, each configuration from the pool is conducted 5 times to reduce randomness.

\subsection{GPCA vs GCN}
\subsubsection{Unsupervised GPCA}
\vspace{-0.05in}

Having proved the mathematical connection between GPCA and GCN, we hypothesize that unsupervised GPCA can generate a comparable representation to (supervised) GCN. 
To test this conjecture, we perform GPCA with different $\alpha$ to obtain node representations and then pass those to a 1- or 2-layer MLP. (We use 2-layer MLP for \arxiv and \pro as these datasets are large.) The result is shown in Table \ref{tb:gpca-gcn}.  For reference, the pool for $\alpha$ is $\{1,5,10,20, 50\}$.

\begin{table}[h]
\vskip -0.15in
\caption{Comparison btwn. Unsupervised GPCA ($\beta=0$) and GCN on 5 datasets, w.r.t. 
	mean accuracy and standard deviation (in parentheses) on test set over 5 different seeds with selected configuration, at which model
achieves best validation accuracy across all HP configurations in Supp. \ref{apdx:hps}. GPCA (ALL $\alpha$) selects best $\alpha$ based on validation, GPCA with specific $\alpha$ uses fixed $\alpha$.}  
\label{tb:gpca-gcn}
\begin{center}
\begin{scriptsize}
\begin{sc}
\begin{tabular}{lccccc}
\toprule
                   & \cora & \seer  & \pub & \arxiv & \pro\\ 
\midrule
\multirow{2}{*}{GCN}  & 80.62  & 71.25   & 78.42  & 70.64  & 77.90   \\ 
                      & (0.90) & (0.05)  & (0.25) & (0.17) & (0.33)  \\ 
\midrule
\multirow{2}{*}{GPCA (all $\alpha$)}  & 81.10 & 71.80  & 78.78 & 71.86   & 79.23   \\ 
                    & (0.00) & (0.75)  & (0.36) & (0.18) & (0.14)  \\ 
\midrule
\multirow{2}{*}{GPCA-$\alpha{=}1$}   & 72.57 &   70.90 &   76.92 & 65.47&  73.65 \\
                    & (0.79) & (0.58) &  (0.30) & (0.26) &  (0.07) \\
\cmidrule{2-6}
\multirow{2}{*}{GPCA-$\alpha{=}5$}   & 80.95 &   71.80 &   \bf{79.40}  & 70.69&  78.66 \\
                    & (0.17) & (0.75) &  (0.29) & (0.11)&   (0.09)\\
\cmidrule{2-6}
\multirow{2}{*}{GPCA-$\alpha{=}10$}  & \bf{82.23} &   71.65 &   78.78 & 71.37&   \bf{79.24}\\
                    & (0.58) & (0.53) &  (0.36) & (0.09) &  (0.09) \\
\cmidrule{2-6}
\multirow{2}{*}{GPCA-$\alpha{=}20$}  & 82.05 &   \bf{72.15} &   78.15 & \bf{71.86}&   79.23\\
                    & (0.54) & (0.47) &  (0.50) & (0.18) &   (0.14) \\
\cmidrule{2-6}
\multirow{2}{*}{GPCA-$\alpha{=}50$}  & 81.10 &   71.50 &   78.00 & 71.48 &  78.92  \\
                    & (0.00) &    (0.32) &    (0.19) & (0.15 &  (0.10) \\
\bottomrule
\end{tabular}
\end{sc}
\end{scriptsize}
\end{center}
\vskip -0.15in
\end{table}

Surprisingly, the parameter-free, single-layer GPCA paired with MLP {performs consistently better} than the end-to-end supervised, multi-layer GCN model across all 5 datasets. By carefully looking at the performance of GPCA with varying $\alpha$, we find that different datasets have different best $\alpha$ but in general a relatively larger $\alpha$ (comparing with graph convolution of GCN that is equivalent to $\alpha=1$) is preferable for all datasets. Larger $\alpha$ implies stronger graph-regularization on the representations. The outstanding performance of simple GPCA empirically confirms that the power of GCN on SSNC problem comes from graph regularization, which 1) questions whether GCN or other GNNs can really learn useful representations for SSNC problem by taking advantage of deep neural networks; and 2) points toward a new direction of studying different graph based regularizations to design novel GNNs with new inductive bias.

\subsubsection{(Semi-)supervised GPCA}
\vspace{-0.05in}

The representations generated by unsupervised GPCA does not use any label information from training data. We have extended GPCA to (semi-)supervised setting 
with an additional HP $\beta \in [0,1]$ 
that trades-off graph-based regularization due to the actual input graph edges versus the ones added through $YY^T$.
  As such, $\beta{=}0$ corresponds to unsupervised GPCA and larger $\beta$ uses more information from training set. 
  The latter raises the potential issue of overfitting that can hurt performance when $\beta$ is too large, or when there is a distribution shift between the training and test sets. For \arxiv and \pro, we empirically observe that $\beta>0$ always hurts performance, possibly because of the distribution difference between training set and test set as described in OGB \cite{hu2020open}. Therefore we only study the effect of $\beta$ on \cora, \seer and \pub. For reference, the pool for $\beta>0$ is $\{0.1,0.2\}$.
  Results in Table \ref{tb:sgpca-gcn} show that supervised GPCA provides a slight gain over unsupervised GPCA across all 3 datasets. 

\begin{table}[h]
\vskip -0.15in
\caption{Comparison btwn. Supervised GPCA ($\beta$$>$$0$), GCN and Unsupervised GPCA on 5 datasets, w.r.t. mean accuracy and standard deviation (in parentheses) on test set over 5 seeds. }
\label{tb:sgpca-gcn}
\begin{center}
\begin{scriptsize}
\begin{sc}
\begin{tabular}{lccccc}
\toprule
                   & \cora & \seer  & \pub \\ 
\midrule
GCN   & 80.62 (0.90) & 71.25 (0.05)   & 78.42 (0.25) \\
\midrule 
Unsupervised GPCA & 81.10 (0.00) & 71.80 (0.75)  & 78.78 (0.36) \\
\midrule 
Supervised GPCA (all $\beta{>}0$) & 81.17 (0.27) & \bf{73.20 (0.71)}   & \bf{79.40 (0.69)} \\
\midrule
Supervised GPCA-$\beta{=}0.1$ & 81.17 (0.27) & 72.07 (0.37) & \bf{79.40 (0.69)}\\
Supervised GPCA-$\beta{=}0.2$ & \bf{81.90 (0.00)} & \bf{73.20 (0.71)} & 78.73 (0.59) \\
\bottomrule
\end{tabular}
\end{sc}
\end{scriptsize}
\end{center}
\vskip -0.15in
\end{table}


\subsection{\method}
\begin{figure}[!t]
\vskip -0.1in
\begin{center}
\centerline{\includegraphics[width=0.8\columnwidth]{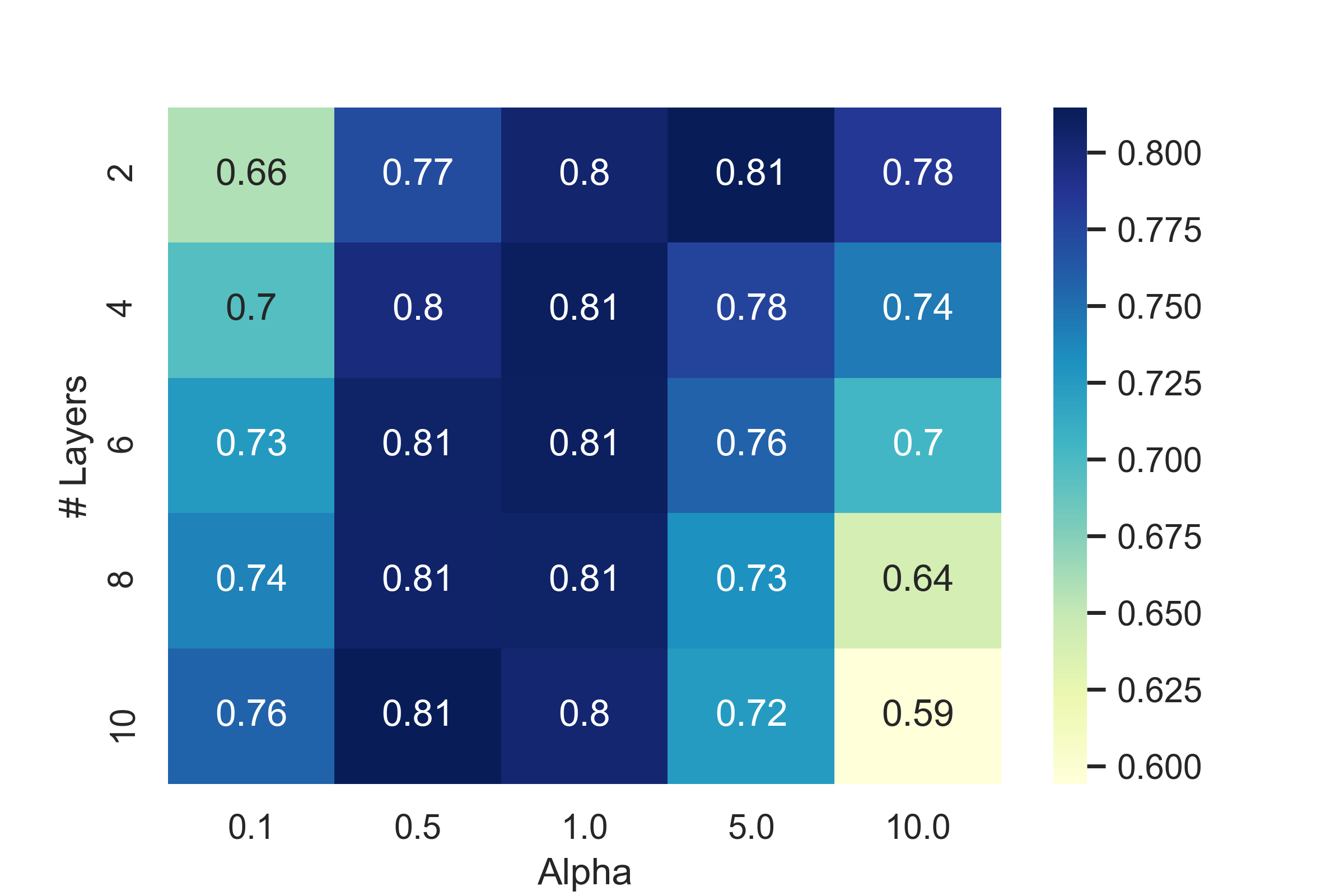}}
\vskip -0.175in
\caption{Performance  of \method (averaged over 5 different seeds) with varying number of layers and varing $\alpha$ on \cora. Similar results hold for \seer and \pub.}
\label{fig:gpcanet}
\end{center}
\vskip -0.1in
\end{figure}

Compared to the single-layer GPCA, 
\method has a deeper architecture like GCN along with nonlinear activation function. Moreover, it employs hyperparameter $\alpha$ at {each} layer to control the degree of graph regularization. As each graph convolution has fixed level of graph regularization, one may hypothesize that increasing number of layers ($L$) of GCN corresponds to increasing degree of graph regularization. We empirically test this hypothesis using \method, by varying both $L$  (2 to 10) and $\alpha$ (0.1 to 10) to show their connection (hidden size is fixed as 128). The result is shown in Figure \ref{fig:gpcanet}. The diagonal pattern empirically suggests that increasing the number of layers has the same effect as increasing graph regularization via $\alpha$. 

The corresponding effect between $\alpha$ and number of layers suggests that we can train a \method with fewer number of layers yet achieve similar regularization by increasing $\alpha$. 
Such a shallow model that otherwise behaves like a deep one has the advantage of less memory requirement and faster training due to fewer parameters.

To this end, we
 train $1$--$3$-layer \method with varying $\alpha$ (see Supp. \ref{apdx:123gpcanet}), and select the best $\alpha$ and number of layers using validation set. We report test set performance in Table \ref{tb:gpcanet}. 
 We do not observe much improvement by \method over GCN on smaller datasets \cora, \seer, \pub, but notable gains on the larger \arxiv and \pro. As such, \method enables shallow model training via tunable hyperparameter $\alpha$, achieving comparable or better performance.


\begin{table}[h]
\vskip -0.15in
\caption{Comparison btwn. $1$--$3$-layer \method and GCN on 5 datasets, w.r.t. mean accuracy and standard deviation (in parentheses) on test set over 5 different seeds.}  
\label{tb:gpcanet}
\begin{center}
\begin{scriptsize}
\begin{sc}
\begin{tabular}{lccccc}
\toprule
                    & \cora & \seer  & \pub & \arxiv & \pro\\ 
\midrule
\multirow{2}{*}{GCN}  & 80.62  & 71.25   & 78.42  & 70.64  & 77.90   \\ 
                      & (0.90) & (0.05)  & (0.25) & (0.17) & (0.33)  \\ 
\midrule
\multirow{2}{*}{\method}  & 80.64 & 71.36 & 78.52 & \bf{72.20} & \bf{80.05}\\
                          & (0.33) & (0.21)& (0.17) & (0.15) & (0.29) \\
\bottomrule
\end{tabular}
\end{sc}
\end{scriptsize}
\end{center}
\vskip -0.15in
\end{table}


\subsection{\method-initialization for GCN}
\vspace{-0.05in}

Finally, we 
evaluate the effectiveness of \method-initialization for GCN in terms of performance 
and robustness under different model sizes, i.e. number of layers $L$. 
For comparison, Glorot et al.'s \yrcite{glorot2010understanding}  Xavier initialization is used to initialize GCN. 

We report the test set performance (averaged over 5 seeds) of the GCN model with the best configuration based on validation data in Table \ref{tab:gpca-init}.
 The results show that \method-initialization tends to outperform the widely-used Xavier initialization, where the improvement 
 grows with increasing number of layers.
 Notably, 
  GCN with \method-initialization exhibits stable performance across all layers.

\begin{table}[!t]
\vskip -0.1in
\caption{Test set performance of GCN with Xaiver initialization versus \method initialization, w.r.t. varying number of layers ($L$) across all datasets. 
	Each reported value is based on the best configuration selected using validation performance.} 
\label{tab:gpca-init}
\begin{center}
\begin{scriptsize}
\begin{sc}
\begin{tabular}{lccccr}
\toprule
Dataset       & $2L$ & $3L$ & $5L$ & $10L$ & $15L$\\
\midrule
\cora~ Xaiver-init     & 80.62 & \bf{80.62} & 79.40  & 76.37 & 66.07 \\
\cora~ \method-init  & \bf{81.67} & 79.50 & \bf{80.90} & \bf{79.82} & \bf{78.00}  \\
\midrule
\seer~ Xaiver-init     & 71.25 & \bf{70.15} &  \bf{71.10} &  61.90 &  57.40  \\
\seer~ \method-init  & \bf{71.27} & 69.27 & 70.15 & \bf{68.67} & \bf{67.87}\\
\midrule
\pub~ Xaiver-init     & \bf{78.42} & \bf{77.90} &  77.07 &  77.00 &  45.80 \\
\pub~ \method-init  & 78.05 & 77.25 & \bf{78.07} & \bf{77.80} & \bf{78.03} \\
\midrule
\arxiv~ Xaiver-init     & 69.61 & \bf{70.64} &  70.33 &  68.32 &  61.68 \\
\arxiv~ \method-init  & \bf{69.76} &  70.72 & \bf{70.52} & \bf{69.77} &  \bf{66.28}\\
\midrule
\pro~ Xaiver-init     & 77.90 & 78.65 & 78.08 & 76.27 & 74.70 \\
\pro~ \method-init  & \bf{78.13} & \bf{78.71} & \bf{78.22} & \bf{77.47} & \bf{75.90} \\
\bottomrule
\end{tabular}
\end{sc}
\end{scriptsize}
\end{center}
\vskip -0.2in
\end{table}
\begin{figure}[!h]
	\begin{center}
		\centerline{\includegraphics[width=\columnwidth]{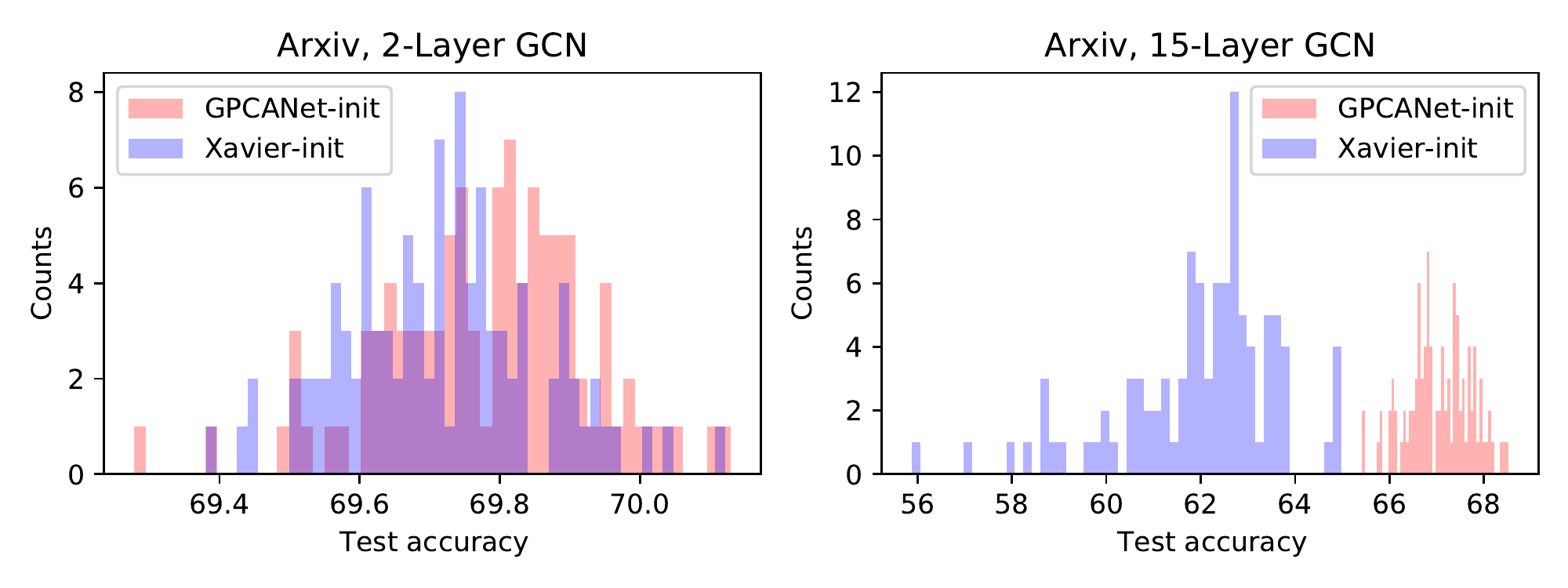}}
			\vskip -0.2in
		\caption{Comparison between Xavier-init and \method-init in terms of test accuracy robustness over 100 seeds on \arxiv. }
		\label{fig:init-robust}
	\end{center}
\vskip -0.1in
\end{figure}


Instead of only looking at the average performance, we further study whether \method-initialization improves the training robustness, by reducing performance variation across different seeds. 
To this end, we first choose the best configuration for each initialization method based on validation performance, and train the GCN model with the chosen configuration using 100 random seeds.

In Figure \ref{fig:init-robust} we present the histogram of test set accuracy over 100 runs for \arxiv. (For results on other datasets, see Supp. \ref{apdx:init-robust}.) 
For both 2-layer and 15-layer GCN, \method-initialization not only outperforms Xavier-initialization w.r.t. average performance, but also in terms of robustness, achieving much lower performance variation and few bad outliers, especially for deeper GCN. 
As such, it acts as a 
good, data-driven prior, facilitating the training of many parameters across layers by identifying a promising region of
the parameter space from which supervised fine-tuning is
initiated.

%

\hide{
\subsubsection{Fewer Training Samples}
For Arxiv and Products, not only dataset size is large but also the size of training set (around 60\% nodes for Arxiv and 8\% for Products). We hypothesize that data-dependent initialization works better when training set is smaller, as bad initialization is harder to escape without enough guidance from labels. We randomly down-sampled the training set of both Arxiv and Products to only $1\%$ of the original training set, and keep validation and test set unchanged. Figure \ref{fig:init-downsample} shows that: 1) surprisingly the test set performance is robust to reducing training set size compared with full-training-set result in Table \ref{tab:gpca-init}; 2) the improvement from \method-initialization seems the same comparing with full-training-set case, however it does help to train deeper model to achieve higher performance.

\begin{figure}[ht]
\vskip 0.2in
\begin{center}
\centerline{\includegraphics[width=\columnwidth]{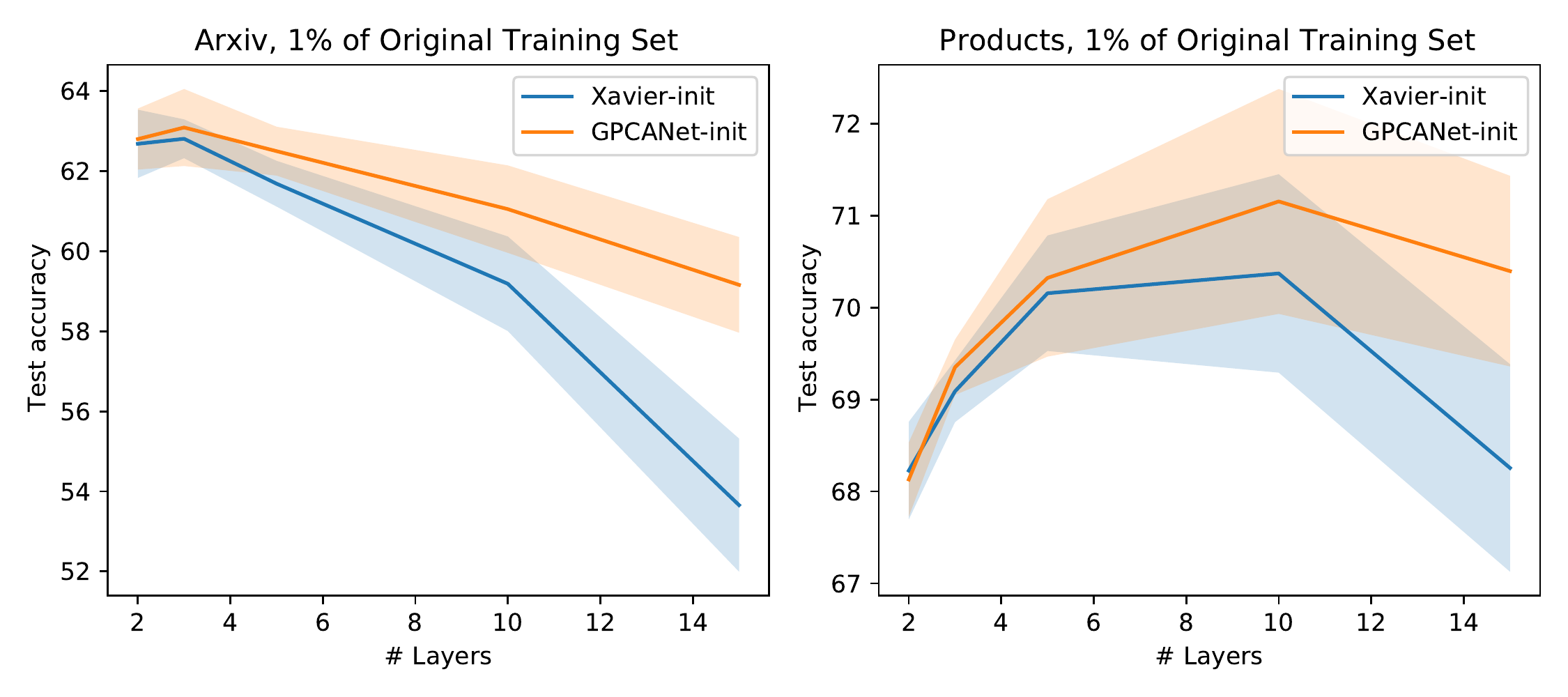}}
\caption{Comparison between Xavier-init and \method-init of GCN on Arxiv and Products, with reduced labeling size. Shaded area represents standard derivation (over 5 runs).}
\label{fig:init-downsample}
\end{center}
\vskip -0.2in
\end{figure}
}

\vspace{-0.05in}
\section{Conclusion}
    \label{sec:end}

\vspace{-0.05in}
In this work we have (1) discovered a mathematical connection between GPCA and GCN's graph convolution; (2) extended GPCA to the (semi-)supervised setting;  (3) proposed \method, by stacking GPCA and nonlinear transformation, which is a generalized GCN model with additional hyperparameters to control the degree of graph regularization, and (4) based on the established connection, proposed the \method-initialization for GCN. 
Accordingly, we designed extensive experiments demonstrating that ($i$) the simple, single-layer GPCA achieves comparable or better performance than GCN, which suggests that GCN's power is driven by graph regularization,
($ii$) \method allows training of shallow models with competitive performance via increasing the degree of graph regularization at each layer, reducing memory and training time cost, and finally ($iii$) \method-initialization acts as a good data-driven prior for GCN training, providing robust performance. 




\clearpage
\bibliography{reference}

\begin{thebibliography}{41}
\providecommand{\natexlab}[1]{#1}
\providecommand{\url}[1]{\texttt{#1}}
\expandafter\ifx\csname urlstyle\endcsname\relax
  \providecommand{\doi}[1]{doi: #1}\else
  \providecommand{\doi}{doi: \begingroup \urlstyle{rm}\Url}\fi

\bibitem[Bengio et~al.(2013)Bengio, Courville, and Vincent]{Bengio13}
Bengio, Y., Courville, A., and Vincent, P.
\newblock Representation learning: A review and new perspectives.
\newblock \emph{IEEE Transactions on Pattern Analysis and Machine
  Intelligence}, 35\penalty0 (8):\penalty0 1798--1828, August 2013.
\newblock ISSN 0162-8828.
\newblock \doi{10.1109/TPAMI.2013.50}.
\newblock URL \url{http://ieeexplore.ieee.org/document/6472238/}.
\newblock Zu bearbeitendes Review.

\bibitem[Chan et~al.(2015)Chan, Jia, Gao, Lu, Zeng, and Ma]{chan2015pcanet}
Chan, T.-H., Jia, K., Gao, S., Lu, J., Zeng, Z., and Ma, Y.
\newblock {PCANet}: A simple deep learning baseline for image classification?
\newblock \emph{IEEE transactions on image processing}, 24\penalty0
  (12):\penalty0 5017--5032, 2015.

\bibitem[Chen et~al.(2020)Chen, Chen, Villar, and Bruna]{chen2020can}
Chen, Z., Chen, L., Villar, S., and Bruna, J.
\newblock Can graph neural networks count substructures?
\newblock \emph{arXiv preprint arXiv:2002.04025}, 2020.

\bibitem[Chiang et~al.(2019)Chiang, Liu, Si, Li, Bengio, and
  Hsieh]{chiang2019cluster}
Chiang, W.-L., Liu, X., Si, S., Li, Y., Bengio, S., and Hsieh, C.-J.
\newblock Cluster-gcn: An efficient algorithm for training deep and large graph
  convolutional networks.
\newblock In \emph{Proceedings of the 25th ACM SIGKDD International Conference
  on Knowledge Discovery \& Data Mining}, pp.\  257--266, 2019.

\bibitem[Errica et~al.(2020)Errica, Podda, Bacciu, and Micheli]{Errica2020A}
Errica, F., Podda, M., Bacciu, D., and Micheli, A.
\newblock A fair comparison of graph neural networks for graph classification.
\newblock In \emph{International Conference on Learning Representations}, 2020.
\newblock URL \url{https://openreview.net/forum?id=HygDF6NFPB}.

\bibitem[Fey \& Lenssen(2019)Fey and Lenssen]{Fey/Lenssen/2019}
Fey, M. and Lenssen, J.~E.
\newblock Fast graph representation learning with {PyTorch Geometric}.
\newblock In \emph{ICLR Workshop on Representation Learning on Graphs and
  Manifolds}, 2019.

\bibitem[Gallagher et~al.(2008)Gallagher, Tong, Eliassi-Rad, and
  Faloutsos]{conf/kdd/GallagherTEF08}
Gallagher, B., Tong, H., Eliassi-Rad, T., and Faloutsos, C.
\newblock Using ghost edges for classification in sparsely labeled networks.
\newblock In \emph{KDD}, pp.\  256--264. ACM, 2008.
\newblock URL
  \url{http://dblp.uni-trier.de/db/conf/kdd/kdd2008.html#GallagherTEF08}.

\bibitem[Garg et~al.(2020)Garg, Jegelka, and Jaakkola]{garg2020generalization}
Garg, V., Jegelka, S., and Jaakkola, T.
\newblock Generalization and representational limits of graph neural networks.
\newblock In \emph{International Conference on Machine Learning}, pp.\
  3419--3430. PMLR, 2020.

\bibitem[Glorot \& Bengio(2010)Glorot and Bengio]{glorot2010understanding}
Glorot, X. and Bengio, Y.
\newblock Understanding the difficulty of training deep feedforward neural
  networks.
\newblock In \emph{Proceedings of the thirteenth international conference on
  artificial intelligence and statistics}, pp.\  249--256, 2010.

\bibitem[Golub \& {Van Loan}(1989)Golub and {Van Loan}]{matrixmethods89}
Golub, G. and {Van Loan}, C.
\newblock \emph{Matrix Computations}.
\newblock Johns Hopkins University Press, 1989.

\bibitem[Hamilton et~al.(2017)Hamilton, Ying, and
  Leskovec]{hamilton2017inductive}
Hamilton, W., Ying, Z., and Leskovec, J.
\newblock Inductive representation learning on large graphs.
\newblock In \emph{Advances in neural information processing systems}, pp.\
  1024--1034, 2017.

\bibitem[He et~al.(2015)He, Zhang, Ren, and Sun]{he2015delving}
He, K., Zhang, X., Ren, S., and Sun, J.
\newblock Delving deep into rectifiers: Surpassing human-level performance on
  imagenet classification.
\newblock In \emph{Proceedings of the IEEE international conference on computer
  vision}, pp.\  1026--1034, 2015.

\bibitem[Hu et~al.(2020)Hu, Fey, Zitnik, Dong, Ren, Liu, Catasta, and
  Leskovec]{hu2020open}
Hu, W., Fey, M., Zitnik, M., Dong, Y., Ren, H., Liu, B., Catasta, M., and
  Leskovec, J.
\newblock Open graph benchmark: Datasets for machine learning on graphs.
\newblock \emph{arXiv preprint arXiv:2005.00687}, 2020.

\bibitem[Ioffe \& Szegedy(2015)Ioffe and Szegedy]{ioffe2015batch}
Ioffe, S. and Szegedy, C.
\newblock Batch normalization: Accelerating deep network training by reducing
  internal covariate shift.
\newblock In \emph{International conference on machine learning}, pp.\
  448--456. PMLR, 2015.

\bibitem[Jiang et~al.(2013)Jiang, Ding, Luo, and Tang]{jiang2013graph}
Jiang, B., Ding, C., Luo, B., and Tang, J.
\newblock Graph-laplacian {PCA}: Closed-form solution and robustness.
\newblock In \emph{Proceedings of the IEEE Conference on Computer Vision and
  Pattern Recognition}, pp.\  3492--3498, 2013.

\bibitem[Kipf \& Welling(2017)Kipf and Welling]{kipf2017semi}
Kipf, T.~N. and Welling, M.
\newblock Semi-supervised classification with graph convolutional networks.
\newblock In \emph{International Conference on Learning Representations
  (ICLR)}, 2017.

\bibitem[Klicpera et~al.(2019)Klicpera, Bojchevski, and
  G{\"u}nnemann]{klicpera_predict_2019}
Klicpera, J., Bojchevski, A., and G{\"u}nnemann, S.
\newblock Predict then propagate: Graph neural networks meet personalized
  pagerank.
\newblock In \emph{International Conference on Learning Representations
  (ICLR)}, 2019.

\bibitem[Kr{\"a}henb{\"u}hl et~al.(2016)Kr{\"a}henb{\"u}hl, Doersch, Donahue,
  and Darrell]{krahenbuhl2015data}
Kr{\"a}henb{\"u}hl, P., Doersch, C., Donahue, J., and Darrell, T.
\newblock Data-dependent initializations of convolutional neural networks.
\newblock In \emph{International Conference on Learning Representations}, 2016.

\bibitem[Krizhevsky et~al.(2012)Krizhevsky, Sutskever, and
  Hinton]{krizhevsky2012imagenet}
Krizhevsky, A., Sutskever, I., and Hinton, G.~E.
\newblock Imagenet classification with deep convolutional neural networks.
\newblock In \emph{Advances in neural information processing systems}, pp.\
  1097--1105, 2012.

\bibitem[Li et~al.(2018)Li, Han, and Wu]{li2018deeper}
Li, Q., Han, Z., and Wu, X.-M.
\newblock Deeper insights into graph convolutional networks for semi-supervised
  learning.
\newblock In \emph{Thirty-Second AAAI Conference on Artificial Intelligence},
  2018.

\bibitem[Liao et~al.(2021)Liao, Urtasun, and Zemel]{liao2021a}
Liao, R., Urtasun, R., and Zemel, R.
\newblock A {\{}pac{\}}-bayesian approach to generalization bounds for graph
  neural networks.
\newblock In \emph{International Conference on Learning Representations}, 2021.
\newblock URL \url{https://openreview.net/forum?id=TR-Nj6nFx42}.

\bibitem[Loukas(2020{\natexlab{a}})]{Loukas2020What}
Loukas, A.
\newblock What graph neural networks cannot learn: depth vs width.
\newblock In \emph{International Conference on Learning Representations},
  2020{\natexlab{a}}.
\newblock URL \url{https://openreview.net/forum?id=B1l2bp4YwS}.

\bibitem[Loukas(2020{\natexlab{b}})]{loukas2020hard}
Loukas, A.
\newblock How hard is to distinguish graphs with graph neural networks?
\newblock Technical report, 2020{\natexlab{b}}.

\bibitem[Low et~al.(2017)Low, Teoh, and Toh]{low2017stacking}
Low, C.-Y., Teoh, A. B.-J., and Toh, K.-A.
\newblock Stacking {PCANet+}: An overly simplified convnets baseline for face
  recognition.
\newblock \emph{IEEE Signal Processing Letters}, 24\penalty0 (11):\penalty0
  1581--1585, 2017.

\bibitem[Mishkin \& Matas(2015)Mishkin and Matas]{mishkin2015all}
Mishkin, D. and Matas, J.
\newblock All you need is a good init.
\newblock \emph{arXiv preprint arXiv:1511.06422}, 2015.

\bibitem[Morris et~al.(2019)Morris, Ritzert, Fey, Hamilton, Lenssen, Rattan,
  and Grohe]{morris2019weisfeiler}
Morris, C., Ritzert, M., Fey, M., Hamilton, W.~L., Lenssen, J.~E., Rattan, G.,
  and Grohe, M.
\newblock Weisfeiler and leman go neural: Higher-order graph neural networks.
\newblock In \emph{Proceedings of the AAAI Conference on Artificial
  Intelligence}, volume~33, pp.\  4602--4609, 2019.

\bibitem[NT \& Maehara(2019)NT and Maehara]{nt2019revisiting}
NT, H. and Maehara, T.
\newblock Revisiting graph neural networks: All we have is low-pass filters.
\newblock \emph{arXiv preprint arXiv:1905.09550}, 2019.

\bibitem[Oono \& Suzuki(2020)Oono and Suzuki]{Oono2020Graph}
Oono, K. and Suzuki, T.
\newblock Graph neural networks exponentially lose expressive power for node
  classification.
\newblock In \emph{International Conference on Learning Representations}, 2020.
\newblock URL \url{https://openreview.net/forum?id=S1ldO2EFPr}.

\bibitem[Pan \& Chen(1999)Pan and Chen]{pan1999complexity}
Pan, V.~Y. and Chen, Z.~Q.
\newblock The complexity of the matrix eigenproblem.
\newblock In \emph{Proceedings of the thirty-first annual ACM symposium on
  Theory of computing}, pp.\  507--516, 1999.

\bibitem[Saxe et~al.(2013)Saxe, McClelland, and Ganguli]{saxe2013exact}
Saxe, A.~M., McClelland, J.~L., and Ganguli, S.
\newblock Exact solutions to the nonlinear dynamics of learning in deep linear
  neural networks.
\newblock \emph{arXiv preprint arXiv:1312.6120}, 2013.

\bibitem[Sen et~al.(2008)Sen, Namata, Bilgic, Getoor, Galligher, and
  Eliassi-Rad]{sen2008collective}
Sen, P., Namata, G., Bilgic, M., Getoor, L., Galligher, B., and Eliassi-Rad, T.
\newblock Collective classification in network data.
\newblock \emph{AI magazine}, 29\penalty0 (3):\penalty0 93--93, 2008.

\bibitem[Seuret et~al.(2017)Seuret, Alberti, Liwicki, and
  Ingold]{seuret2017pca}
Seuret, M., Alberti, M., Liwicki, M., and Ingold, R.
\newblock Pca-initialized deep neural networks applied to document image
  analysis.
\newblock In \emph{2017 14th IAPR international conference on document analysis
  and recognition (ICDAR)}, volume~1, pp.\  877--882. IEEE, 2017.

\bibitem[Shahid et~al.(2016)Shahid, Perraudin, Kalofolias, Puy, and
  Vandergheynst]{shahid2016fast}
Shahid, N., Perraudin, N., Kalofolias, V., Puy, G., and Vandergheynst, P.
\newblock Fast robust {PCA} on graphs.
\newblock \emph{IEEE Journal of Selected Topics in Signal Processing},
  10\penalty0 (4):\penalty0 740--756, 2016.

\bibitem[Srinivasan \& Ribeiro(2020)Srinivasan and Ribeiro]{Srinivasan2020On}
Srinivasan, B. and Ribeiro, B.
\newblock On the equivalence between positional node embeddings and structural
  graph representations.
\newblock In \emph{International Conference on Learning Representations}, 2020.
\newblock URL \url{https://openreview.net/forum?id=SJxzFySKwH}.

\bibitem[Veli{\v{c}}kovi{\'{c}} et~al.(2018)Veli{\v{c}}kovi{\'{c}}, Cucurull,
  Casanova, Romero, Li{\`{o}}, and Bengio]{velickovic2018graph}
Veli{\v{c}}kovi{\'{c}}, P., Cucurull, G., Casanova, A., Romero, A., Li{\`{o}},
  P., and Bengio, Y.
\newblock {Graph Attention Networks}.
\newblock In \emph{International Conference on Learning Representations}, 2018.

\bibitem[Veličković et~al.(2019)Veličković, Fedus, Hamilton, Liò, Bengio,
  and Hjelm]{velickovic2018deep}
Veličković, P., Fedus, W., Hamilton, W.~L., Liò, P., Bengio, Y., and Hjelm,
  R.~D.
\newblock Deep graph infomax.
\newblock In \emph{International Conference on Learning Representations}, 2019.
\newblock URL \url{https://openreview.net/forum?id=rklz9iAcKQ}.

\bibitem[Wagner et~al.(2013)Wagner, Thom, Schweiger, Palm, and
  Rothermel]{wagner2013learning}
Wagner, R., Thom, M., Schweiger, R., Palm, G., and Rothermel, A.
\newblock Learning convolutional neural networks from few samples.
\newblock In \emph{The 2013 International Joint Conference on Neural Networks
  (IJCNN)}, pp.\  1--7. IEEE, 2013.

\bibitem[Wu et~al.(2019)Wu, Souza, Zhang, Fifty, Yu, and
  Weinberger]{wu2019simplifying}
Wu, F., Souza, A., Zhang, T., Fifty, C., Yu, T., and Weinberger, K.
\newblock Simplifying graph convolutional networks.
\newblock In \emph{International Conference on Machine Learning}, pp.\
  6861--6871, 2019.

\bibitem[Xu et~al.(2019)Xu, Hu, Leskovec, and Jegelka]{xu2018how}
Xu, K., Hu, W., Leskovec, J., and Jegelka, S.
\newblock How powerful are graph neural networks?
\newblock In \emph{International Conference on Learning Representations}, 2019.
\newblock URL \url{https://openreview.net/forum?id=ryGs6iA5Km}.

\bibitem[Zhang \& Zhao(2012)Zhang and Zhao]{zhang2012low}
Zhang, Z. and Zhao, K.
\newblock Low-rank matrix approximation with manifold regularization.
\newblock \emph{IEEE transactions on pattern analysis and machine
  intelligence}, 35\penalty0 (7):\penalty0 1717--1729, 2012.

\bibitem[Zhao \& Akoglu(2020)Zhao and Akoglu]{Zhao2020PairNorm:}
Zhao, L. and Akoglu, L.
\newblock Pairnorm: Tackling oversmoothing in gnns.
\newblock In \emph{International Conference on Learning Representations}, 2020.
\newblock URL \url{https://openreview.net/forum?id=rkecl1rtwB}.

\end{thebibliography}
\bibliographystyle{icml2021}



\clearpage
\appendix
\section{Supplementary Material}
\subsection{Derivation from Eq. \eqref{eq:gpcaa} to Eq. \eqref{eq:gpca-eig}}
\label{derivation:trace}

For this part only, let $A=(I+\alpha \tilde{L})^{-1}$ to simplify the notation. We can show that \eqref{eq:objplug} is equivalent to
    \begin{align}
 & \min_{W, W^TW= I}  \quad  \Tr(XX^T) - 2\Tr(AXWW^TX^T) \nonumber \\ 
 & + \Tr (AXWW^TWW^TX^TA) + \alpha \Tr (  W^T X^T A \tilde{L} A XW )\nonumber \\  
 \equiv & \max_{W, W^TW= I}  \quad   2\Tr(AXWW^TX^T) - \Tr (AXWW^TX^TA) \nonumber \\ 
 &  \quad \quad \quad \quad \quad  - \alpha \Tr (  W^T X^T A \tilde{L} A XW )
 \end{align}
 Using the cyclic property of (Tr)ace, we can write
 \begin{align}
 & \max_{W, W^TW= I}  \quad   2\Tr(W^TX^TAXW) - \Tr (W^TX^TAAXW) \nonumber \\ 
 &  \quad \quad \quad \quad \quad - \alpha \Tr (  W^T X^T A \tilde{L} A XW ) \nonumber \\ 
 & \max_{W, W^TW= I}  \quad   \Tr \big[ W^TX^T (2A - AA - A (\alpha \tilde{L}) A ) XW  \big] \nonumber \\ 
 & \max_{W, W^TW= I}  \quad   \Tr \big[ W^TX^T \big(A  + \{I - A (I+\alpha \tilde{L})  \} A  \big) XW  \big]\nonumber \\
 & \max_{W, W^TW= I}  \quad   \Tr \big[ W^TX^T (I+\alpha \tilde{L})^{-1} XW  \big] 
 \end{align}
 where the objective simplifies upon replacing $A$ with $(I+\alpha \tilde{L})^{-1}$.

\subsection{Derivation from Eq. \eqref{eq:corr} to Eq. \eqref{eq:gpca-eig}}
\label{derivation:var}

\begin{align}
\max_{\bz} \quad \; &  \big[\text{corr}(Y, \bz)  \big]^T \big[\text{corr}(Y, \bz)  \big]  \text{var}(\bz) \nonumber\\
\equiv \max_{\bz} \quad \;&  \text{var}(Y)   \big[\text{corr}(Y, \bz)  \big]^T \big[\text{corr}(Y, \bz)  \big]  \text{var}(\bz) \label{eq:addy} \\
\equiv \max_{\bz} \quad \;&   
\big[  \text{cov}(Y, \bz) \big]^T \big[ \text{cov}(Y, \bz)  \big]
\; \\ & \text{ where }
\text{cov}(Y, \bz) =   \sqrt{\text{var}(Y)} \text{corr}(Y, \bz)   \sqrt{\text{var}(\bz)} \nonumber \\
\equiv \max_{\bz}\quad  \;&  \big[ Y^T \bz\big]^T  \big[ Y^T \bz\big] \\
\equiv \max_{\bz} \quad \;&  \bz^T Y Y^T \bz
\end{align}

Note that in \eqref{eq:addy} we added the term $\text{var}(Y)$ without affecting the optimization problem which is w.r.t. $\bz$.

\subsection{Dataset Statistics}\label{apdx:data}

\begin{table}[h!]
\caption{Statistics of used datasets.}
\label{tab:datasets}
\begin{center}
\begin{scriptsize}
\begin{sc}
\begin{tabular}{lccccc}
\toprule
Dataset    & \#Nodes & \#Edges & \#Feat. & \#Cls. & Train/Vali./Test \\
\midrule
Cora       &  2,708   &   5,429& 1,433& 7& 5.2\%/18.5\%/36.9\%\\
CiteSeer   &  3,327   &   4,732& 3,703& 6& 3.6\%/15\%/30\%\\
PubMed     & 19,717   &  44,338&  500& 3& 0.3\%/2.5\%/5\%\\
Arxiv      & 169,343  &1,166,243 & 128&40&54\%/18\%/28\%\\
Products   &2,449,029 & 61,859,140& 100 &47 & 8\%/2\%/90\%\\
\bottomrule
\end{tabular}
\end{sc}
\end{scriptsize}
\end{center}
\vskip -0.1in
\end{table}

Datasets used in the experiments are presented in Table \ref{tab:datasets}. 
Cora, CiteSeer, and PubMed can be downloaded in Pytorch Geometric Library \cite{Fey/Lenssen/2019}.
Arxiv and Products can be accessed in \url{https://ogb.stanford.edu/}.

\subsection{Hyperparameter Configurations}\label{apdx:hps}
We setup hyperparameters pool for each dataset, presented in Table \ref{tab:hps}. All methods use the same pool. 
The only exception is GPCA, as GPCA is just a 1-layer shallow model which can be trained with lager learning rate, we use 0.1 learning rate for it on all datasets. 

\begin{table}[h!]
\caption{Hyperparameters pool for each dataset, includes learning rate (LR), weight decay (WD), 
number of layers (\#Layers), hidden size, dropout, $\alpha$, and $\beta$. For \arxiv and \pro, weight decay is set as 0 because the dataset is large and no overfit happened. Same reason for choosing smaller dropout rate for them.  }
\label{tab:hps}
\begin{flushleft}
\begin{scriptsize}
\begin{sc}
\begin{tabular}{lccccccc}
\toprule
Dataset    &  LR &  WD & \#Layers & Hidden \\
\midrule
\cora       & 0.001& [0.0005, 0.005, 0.05]& [2, 3, 5, 10, 15]& [128, 256] \\
\seer   & 0.001& [0.0005, 0.005, 0.05]& [2, 3, 5, 10, 15]& [128, 256] \\
\pub     & 0.001& [0.0005, 0.005, 0.05]& [2, 3, 5, 10, 15]& [128, 256] \\
\arxiv      & 0.005& 0& [2, 3, 5, 10, 15]& [128, 256] \\
\pro   & 0.001& 0& [2, 3, 5, 10, 15]& [128, 256]\\
\bottomrule
\end{tabular}
\begin{tabular}{lccccccc}
\toprule
Dataset      & Dropout & $\alpha$ & $\beta$ \\
\midrule
\cora       & [0, 0.5] & [1, 5, 10, 20, 50]& [0, 0.1, 0.2] \\
\seer       & [0, 0.5] & [1, 5, 10, 20, 50]& [0, 0.1, 0.2] \\
\pub     & [0, 0.5] & [1, 5, 10, 20, 50]& [0, 0.1, 0.2] \\
\arxiv       & [0, 0.2] & [1, 5, 10, 20, 50]& 0 \\
\pro    & [0, 0.1] & [1, 5, 10, 20, 50]& 0\\
\bottomrule
\end{tabular}
\end{sc}
\end{scriptsize}
\end{flushleft}
\vskip -0.1in
\end{table}

\subsection{Configurations for Experiments of $1{\sim}3$-Layer \method} \label{apdx:123gpcanet}
The goal is train a shallow \method with tunable $\alpha$ ($\beta{=}0$ is used), we setup different $\alpha$ pool for different number of layers, because the effect of increasing $\alpha$ is the same to increasing number of layers (shown in Figure \ref{fig:gpcanet}). We report the pool for $\alpha$ for each layer in Table \ref{tab:hps-gpcanet}. For other parameters we use the same setting mentioned in Table \ref{tab:hps}.

\begin{table}[h!]
\caption{Pool of $\alpha$ for $1{\sim}3$-layer \method, same across all datasets.}
\label{tab:hps-gpcanet}
\begin{center}
\begin{small}
\begin{sc}
\begin{tabular}{lc}
\toprule
\# Layers & pool of $\alpha$  \\
\midrule
1-layer     & [10, 20, 30] \\
2-layer     & [3, 5, 10]\\
3-layer     & [1, 2, 3, 5]\\
\bottomrule
\end{tabular}
\end{sc}
\end{small}
\end{center}
\vskip -0.1in
\end{table}

\subsection{\method-Init's Robustness for Additional Datasets}\label{apdx:init-robust}
\begin{figure}[!h]
    \begin{center}
        \centerline{\includegraphics[width=\columnwidth]{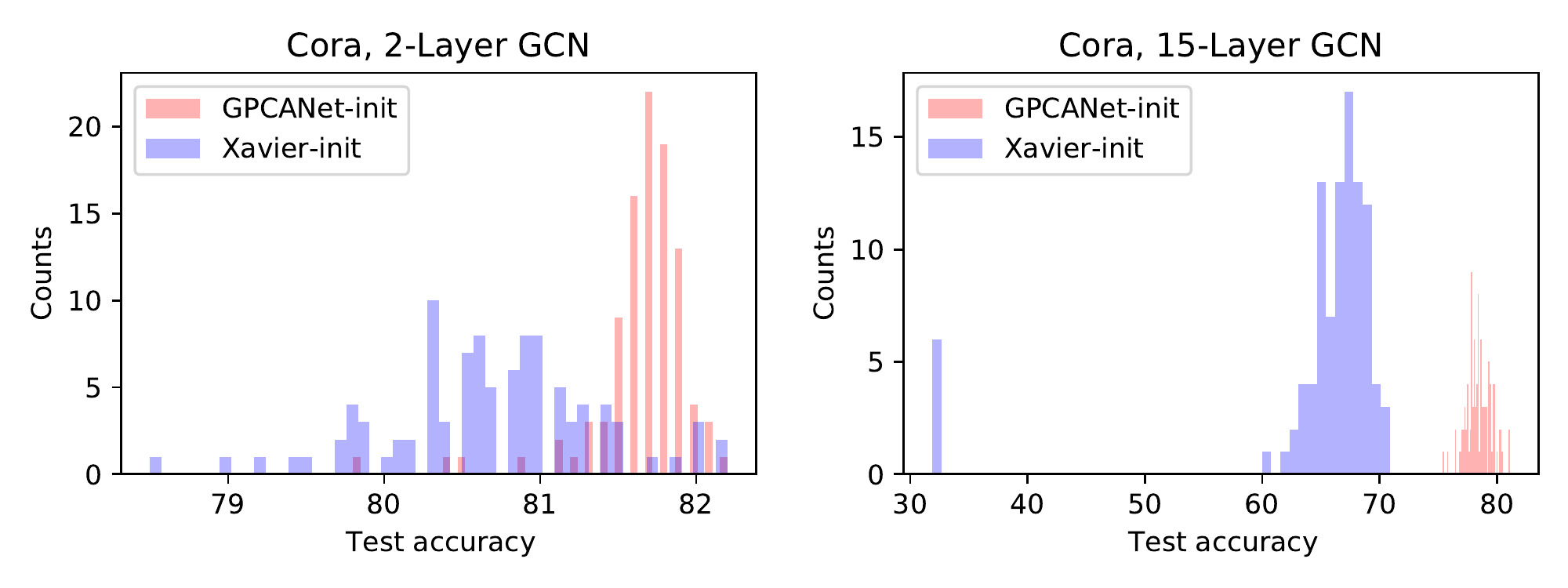}}
            \vskip -0.2in
        \caption{Comparison between Xavier-init and \method-init in terms of test accuracy robustness over 100 seeds on \cora. }
        \label{fig:init-robust-cora}
    \end{center}
\vskip -0.1in
\end{figure}

\begin{figure}[!h]
    \begin{center}
        \centerline{\includegraphics[width=\columnwidth]{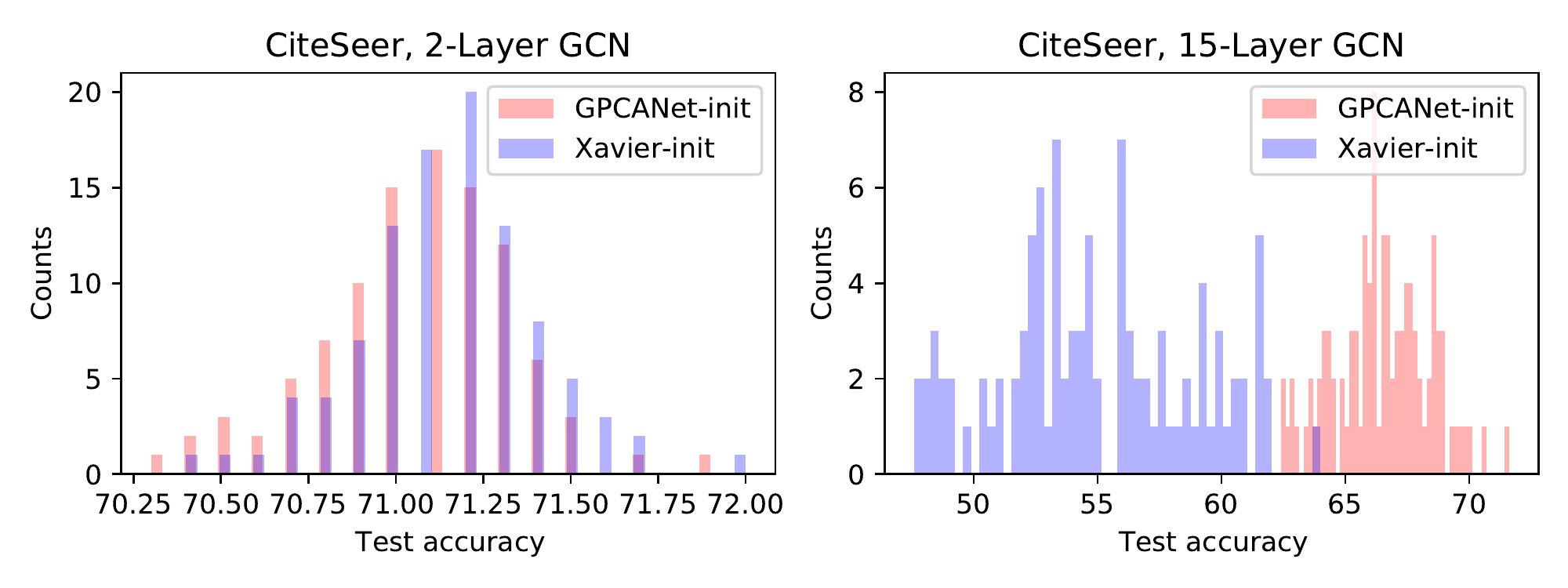}}
            \vskip -0.2in
        \caption{Comparison between Xavier-init and \method-init in terms of test accuracy robustness over 100 seeds on \seer. }
        \label{fig:init-robust-citeseer}
    \end{center}
\vskip -0.1in
\end{figure}

\begin{figure}[!h]
    \begin{center}
        \centerline{\includegraphics[width=\columnwidth]{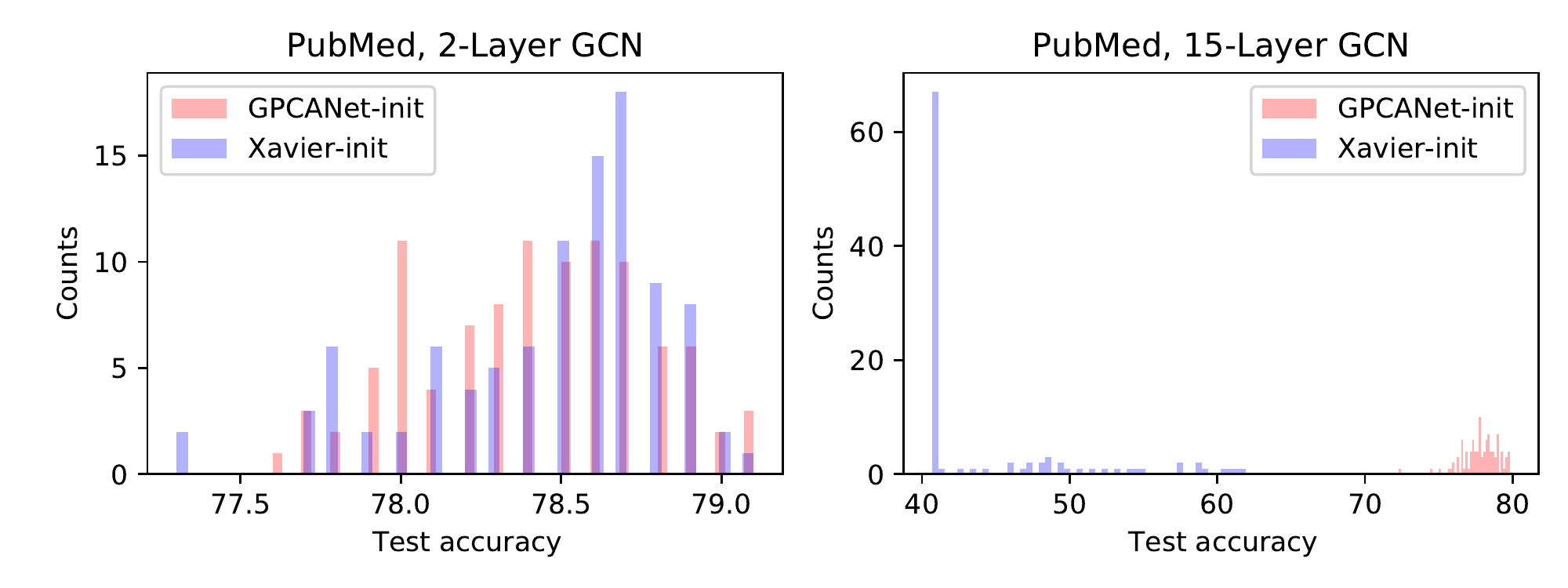}}
            \vskip -0.2in
        \caption{Comparison between Xavier-init and \method-init in terms of test accuracy robustness over 100 seeds on \pub. }
        \label{fig:init-robust-pubmed}
    \end{center}
\vskip -0.1in
\end{figure}

Histogram of test set accuracy over 100 runs for GCN initialized by Xavier-initialization and \method-initialization in \cora (Figure \ref{fig:init-robust-cora}), \seer (Figure \ref{fig:init-robust-citeseer}), and \pub (Figure \ref{fig:init-robust-pubmed}). We have ignored \pro as it takes too long to run 100 times, but the result should be similar. 

\end{document}